\documentclass[conference]{IEEEtran}
\IEEEoverridecommandlockouts
\usepackage{cite}
\usepackage{amsmath,amssymb,amsfonts}
\usepackage{algorithmic}
\usepackage{graphicx}
\usepackage{textcomp}
\usepackage{xcolor}
\usepackage{hyperref}
\usepackage{subfigure}

\def\BibTeX{{\rm B\kern-.05em{\sc i\kern-.025em b}\kern-.08em
  T\kern-.1667em\lower.7ex\hbox{E}\kern-.125emX}}
  
\makeatletter
\newcommand{\linebreakand}{%
 \end{@IEEEauthorhalign}
 \hfill\mbox{}\par
 \mbox{}\hfill\begin{@IEEEauthorhalign}
}
\makeatother

\begin{document}

\title{COPER: Continuous Patient State Perceiver\\
\thanks{
This work was supported in part by the National Institute for Health Research (NIHR) Oxford Biomedical Research Centre (BRC) and in part by InnoHK Project Programme 3.2: Human Intelligence and AI Integration (HIAI) for the Prediction and Intervention of CVDs: Warning System at Hong Kong Centre for Cerebro-cardiovascular Health Engineering (COCHE).

Vinod Kumar Chauhan is supported by a Medical Research Council (MRC) Research Grant (MR/W01761X/1).

The views expressed are those of the authors and not necessarily those of the NHS, the NIHR, the Department of Health, the InnoHK – ITC or the MRC.
}
}

\author{\IEEEauthorblockN{Vinod Kumar Chauhan}
\IEEEauthorblockA{\textit{Institute of Biomedical Engineering,} \\
\textit{University of Oxford}\\
Oxford, UK \\
Email: \href{mailto:vinod.kumar@eng.ox.ac.uk}{vinod.kumar@eng.ox.ac.uk}}
\and
\IEEEauthorblockN{Anshul Thakur}
\IEEEauthorblockA{\textit{Institute of Biomedical Engineering,} \\
\textit{University of Oxford}\\
Oxford, UK \\
Email: \href{mailto:anshul.thakur@eng.ox.ac.uk}{anshul.thakur@eng.ox.ac.uk}}
\and
\IEEEauthorblockN{Odhran O'Donoghue}
\IEEEauthorblockA{\textit{Institute of Biomedical Engineering,} \\
\textit{University of Oxford}\\
Oxford, UK \\
Email: \href{mailto:odhran.odonoghue@eng.ox.ac.uk}{odhran.odonoghue@eng.ox.ac.uk}}
\linebreakand 

\IEEEauthorblockN{David Andrew Clifton}
\IEEEauthorblockA{\textit{Institute of Biomedical Engineering,} \\
\textit{University of Oxford}\\
Oxford, UK \\
Email: \href{mailto:davidc@robots.ox.ac.uk}{davidc@robots.ox.ac.uk}}
}

\maketitle

\begin{abstract}
In electronic health records (EHRs), irregular time-series (ITS) occur naturally due to patient health dynamics, reflected by irregular hospital visits, diseases/conditions and the necessity to measure different vitals signs at each visit etc. ITS present challenges in training machine learning algorithms which mostly are built on assumption of coherent fixed dimensional feature space.
In this paper, we propose a novel COntinuous patient state PERceiver model, called COPER, to cope with ITS in EHRs. COPER uses Perceiver model and the concept of neural ordinary differential equations (ODEs) to learn the continuous time dynamics of patient state, i.e., continuity of input space and continuity of output space. The neural ODEs help COPER to generate regular time-series to feed to Perceiver model which has the capability to handle multi-modality large-scale inputs.
To evaluate the performance of the proposed model, we use in-hospital mortality prediction task on MIMIC-III dataset and carefully design experiments to study irregularity. The results are compared with the baselines which prove the efficacy of the proposed model.
\end{abstract}


\section{Introduction}
\label{sec_intro}
Irregular time-series (ITS) are the results of uneven time intervals between different measurements, which lead to missing values \cite{Lee2022} and can also create sparsity. In healthcare, ITS occur due to several reasons such as irregularity in the patient visits itself, dependence of physiological measurements at each visit on the patient health status and availability of staff to take measurements etc. \cite{Sharma2022}. ITS are prevalent in primary care as well as secondary care, including critical care, e.g., the MIMIC-III dataset has a general missing rate of 85\% rising to over 90\% for hourly sampling rate data \cite{sun2020review}.

Use of electronic health records (EHRs) have greatly helped to develop machine learning (ML) models to guide clinical decision making, reduce the workload on an already burdened system and increase efficiency of healthcare resources \cite{shickel2017deep,thakur2021dynamic}. However, ML models are mostly based on the assumption of coherent fixed-dimensional feature space, which is invalidated by irregularity in the EHRs, and it is challenging to train ML models without affecting performance \cite{marlin2012unsupervised}. Thus, it is crucial for several reasons, including diagnosis, prognosis, treatment and resource management, to develop techniques for the accurate estimation of missing time-steps.

Due to the wide prevalence of ITS in EHRs and its importance, it have received increasing attention and there has been extensive research in recent years to deal with irregularity, e.g., \cite{rubanova2019latent,shukla2021multitime,Sharma2022,Lee2022}. A wide variety of techniques have been developed to model ITS, e.g., simple statistical techniques for replacement, such as mean value, interpolation, imputation and matrix completion-based techniques \cite{yoon2018estimating} to modification of recurrent neural networks \cite{tan2020data}, neural ODEs \cite{rubanova2019latent}, neural processes \cite{Sharma2022} and attention based techniques \cite{Lee2022} etc.

The conventional simple statistical techniques for replacement, such as mean, median, zero and carry forward are biased and make strong assumptions about the data generating process, and results in loss of performance in downstream prediction tasks \cite{little2019statistical}. Many other modern techniques for modelling fail to capture feature-correlations in time-series \cite{tan2020data}, separate the modelling of missingness from the downstream task, and fail to learn the missingness pattern, or are not adequately efficient to handle large-scale inputs and multi-modality of the data \cite{yoon2018estimating} etc. Moreover, there are few techniques that can handle completely missing time steps, such as in \cite{rubanova2019latent,chen2018neural} but others that can only handle partially missing values, such as in \cite{Lee2022}. 

In this paper, we propose a novel COntinuous patient state PERceiver model (COPER), for end-to-end learning from the ITS in EHRs, which can handle completely missing time steps. COPER uses neural ODEs to obtain continuous time dynamics of patient state from which completely missing steps can be calculated to feed to the Perceiver model \cite{jaegle2021Perceiver}. The Perceiver model learns the temporal dynamics of the time-series and has the capability to deal with large-scale multi-modality inputs.

The first novelty of our work is in learning continuous patient states using embedding and neural ODEs. This is different from the literature where ODEs are used for the obtaining the continuity of the latent representations \cite{chen2018neural,rubanova2019latent}. The continuous patient states could be helpful in a wide range of tasks including diagnosis, prognosis and treatment and in disease progression modelling. The second novelty of our work, to the best of our knowledge, is in adapting the Perceiver model \cite{jaegle2021Perceiver} to time-series data. Perceiver models are recent advancements in transformer models \cite{vaswani2017attention} and are capable of working with large-scale multi-modality inputs. This could be very helpful in healthcare as EHRs represent big data about the patients, and taking into consideration the complete and long trajectory of patient states can yield good results for patient care. Similar to \cite{rubanova2019latent}, we also use neural ODEs for the continuity of latent representations of the Perceiver model. Thus, COPER has a double engine -- in terms of continuity in input and output -- enabling it to cope well with the ITS in EHRs. We present the results of an in-hospital mortality task using the MIMIC-III dataset to demonstrate the utility of the proposed model.

% The rest of the paper is organized as: Section~\ref{sec_literature} presents background and some related work, Section~\ref{sec_methodology} describes the proposed model and Section~\ref{sec_results} presents experiments. Finally, the concluding remarks and future scope are discussed in Section~\ref{sec_conclusion}.

\section{Background}
\label{sec_literature}

\subsection{Perceiver}
\label{subsec_Perceiver_lit}
% Transformer \cite{vaswani2017attention} based models have been successful across different domains with different modalities, including time-series in healthcare \cite{song2018attend}. However, the main limitation of these models is their quadratic dependence on the input size, resulting into large computational complexity to deal with the long context inputs which limits their applicability to such problems which are quite normal in healthcare time-series data \cite{rocheteau2021}.

Perceiver \cite{jaegle2021Perceiver} based models are recent advancements to transformer \cite{vaswani2017attention} models which solve the issue of quadratic dependence of transformers on the input by introducing cross-attentions of inputs with a learnable tight latent vector. The cross-attention distils the high dimensional input to low a dimensional latent which is followed by self-attentions on the latent vector, as given below. Suppose, $X_c \in \mathbb{R}^{n \times t' \times e}$ is a time-series data with $n$ patients, $t'$ time steps each represented with an embedding of size $e$, and $Z \in \mathbb{R}^{n \times t'' \times d}$ is a latent vector (of size $t'' \times d$ but repeated to match batch dimension) with $t''$ number of latents ($t'' << t'$) and $d$ is latent dimension, then cross-attention and self-attention operations of the Perceiver can be represented as:
\begin{equation}
  \label{eq_per1}
  Z = f_{\textsc{cross-attend}}(X_c, Z),
\end{equation}
\begin{equation}
  \label{eq_per2}
  Z = f_{\textsc{self-attend}}(Z),
\end{equation}
where cross-attention is calculated using a scaled dot-product attention \cite{vaswani2017attention} as $\sigma\left(Z {X_c}^{\top}/\sqrt{d_k}\right) {X_c}$, and $\sigma$ is softmax activation function, $d_k$ is the dimension of the key vector. Self-attention is also calculated similarly but with same input as query, key and value. In Perceiver \cite{jaegle2021Perceiver}, each cross-attention is followed by more than one self-attention layers which forms one block. There are several blocks of cross- and self-attention layers, each of which take the same input. The Perceiver based models are multi-modality models which can take large inputs and capture long contexts. There have been several other improvements to Perceiver \cite{jaegle2021Perceiver}, e.g., Perceiver IO \cite{jaegle2021PerceiverIO} generalizes the outputs by introducing cross-attentions at the output similar to at input, and Perceiver AR \cite{hawthorne2022general} extends Perceivers to auto-regressive modelling by introducing masking for attentions etc.

\subsection{Neural Ordinary Differential Equations}
\label{subsec_ode}
Neural ordinary differential equations (ODEs) \cite{chen2018neural} are also a recently developed class of neural networks which help to capture the dynamics of a hidden state using ODEs, resulting into a continuous state space. Neural ODEs consist of a neural network which outputs derivative of the hidden state. The derivative of hidden state is fed to a black-box ODE solver which allows to calculate the hidden state at any time step, resulting in a continuous space, as given below.
\begin{equation}
  \label{eq_node1}
  \dfrac{dZ}{dt} = f_\theta \left(Z(t), t \right),
\end{equation}
\begin{equation}
  \label{eq_node2}
  Z_0,...,Z_N = \textsc{ODESolver}\left(f_\theta, Z_0, \left(t_0,...,t_N \right) \right),
\end{equation}
where $Z$ is hidden state, $f_\theta$ is a neural network which parameterizes the derivative of hidden state, i.e., continuous time dynamics. The ODESolver takes the derivative and initial hidden state $Z_0$ and calculates the hidden state at $\left(t_0,...,t_N \right)$.

The concept of neural ODEs is extended to different domains for solving different problems, including solving the irregular time-series problem. \cite{chen2018neural} discussed a case to fit and extrapolate irregular time-series, which was then extended by \cite{rubanova2019latent} to handle ITS by modelling the hidden state dynamics of RNNs using neural ODEs. Similarly, there are other such extensions, such as those in \cite{kidger2020neural} to handle ITS.

% \subsection{Irregular Time-series}
% \label{subsec_timeseries}
% The prevalence of irregular time-series in healthcare has recently attracted increasing attention and a wide range of techniques have been developed to deal with irregularity \cite{rubanova2019latent,shukla2021multitime,Sharma2022,Lee2022}. Some deal with irregularity at the preprocessing time, e.g, Gaussian process regression, interpolation, resampling, autoregression and replacement based on statistical values, like mean, median or mode etc., and some deal in the end-to-end learning process, e.g., adaptation of recurrent networks, like including missing indicators, decay factors and ODE-RNN etc. (refer to \cite{weerakody2021review} for review of such techniques). Based on technology, recently, recurrent networks and attention based models are popular, and later has an advantage of processing in parallel over the former \cite{Lee2022}. ITS techniques depend on the representations used which can be primarily categorised as series, vector or set representations \cite{shukla2020survey}.

\section{Methodology}
\label{sec_methodology}
\begin{figure}[t]
  \centering
  \includegraphics[scale=0.7]{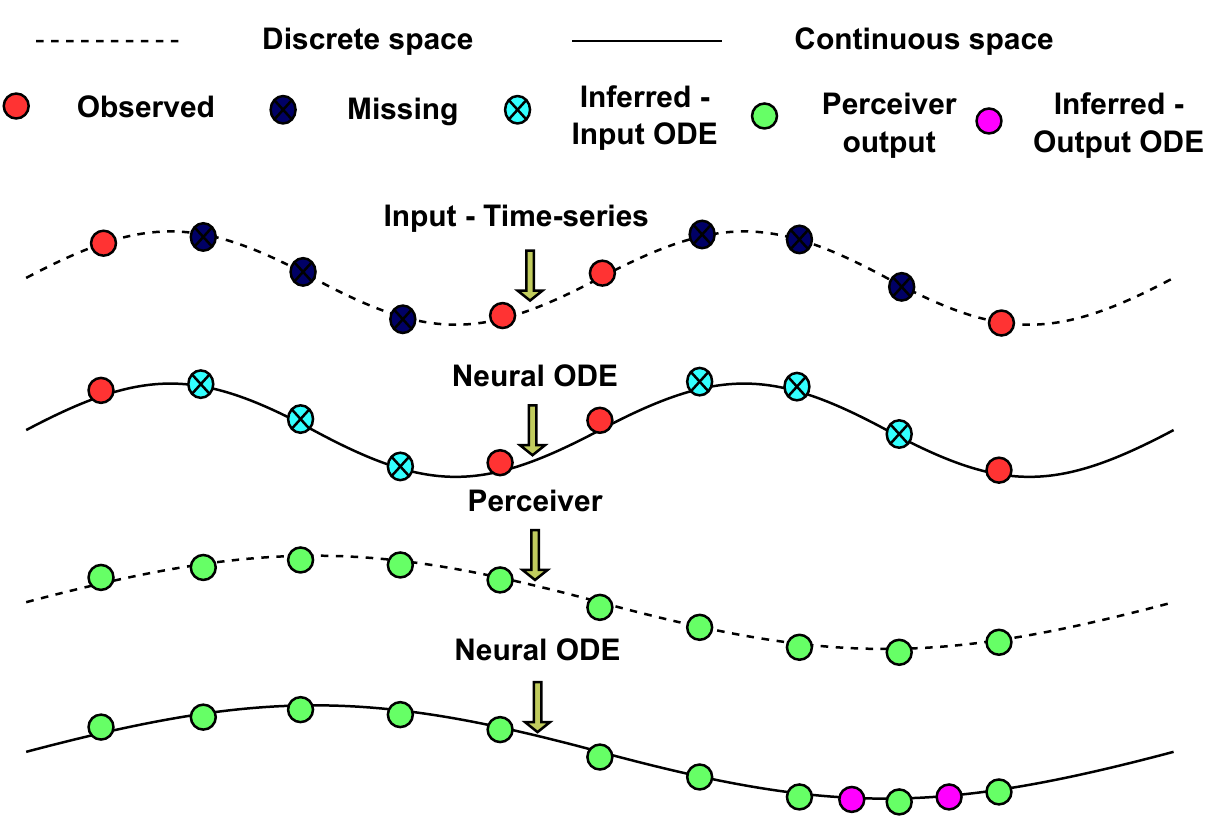}
  \caption{Dynamics of COPER: Neural ODE generates a continuous space from the ITS with missing steps, from which missing observations can be calculated and fed to Perceiver. The discrete latent states generated by Perceiver are again passed to another neural ODE that makes the output space continuous allowing COPER to predict patient condition at any time point.}
  \label{fig_model}
\end{figure}

COPER has the best of neural ODEs and Perceiver networks to handle the ITS in EHRs, and can be applied to different tasks; here, we apply COPER to an in-hospital mortality prediction task. The overall working of COPER is represented in Fig.~\ref{fig_model} using a univariate ITS. It consists of two neural ODEs, one each for the continuity of the input and the output, one Perceiver to learn the hidden time-series representations, and two multi-layer perceptrons (MLPs); one for learning embedding for patient state at each time step and one for the classification task, as discussed below.

\subsection{Embedding}
\label{subsec_embedding}
Embeddings are representations of the input and can be used to adjust the size of the input. Here, the embedding layer is implemented with an MLP with single linear layer of 32 neurons which takes input as a time step (which is a vector of all features/variables at the given time) and produces embedding for the patient state at that point of time, as represented below:
\begin{equation}
  \label{eq_emb}
  X_e = f_{\textsc{emb}}(X),
\end{equation}
where $X \in \mathbb{R}^{n\times t \times i}$ is input ITS data and $X_e \in \mathbb{R}^{n\times t \times e}$ is a learned embedding for $n$ patients, $t$ time steps in the sequence, $i$ input features/variables and $e$ is the size of the embedding.

\subsection{Input Neural ODE}
\label{subsec_ode_in}
A neural ODE, immediately after the embedding layer, helps to make the input space continuous and learns the patient state dynamics from the embedding. This is expected to be less complex to learn compared with learning the hidden state dynamics of recurrent networks in latent ODE \cite{rubanova2019latent} because patient states, mostly, change slowly from one time step to another. This layer could be helpful to generate regular time-series, to work with different ML algorithms and can be independently plugged to other models, and further studied for disease progression modelling.
\begin{equation}
  \label{eq_ode1}
  X_c = f_{\textsc{ode-in}}(X_e),
\end{equation}
where $X_c \in \mathbb{R}^{n \times t' \times e}$ is a regular time-series generated from embedding of ITS and $t'$ is the sequence length.

\subsection{Perceiver}
\label{subsec_Perceiver}
Perceivers are recent advancements to Transformers \cite{vaswani2017attention}, as discussed in \ref{subsec_Perceiver_lit}. In our architecture, we have used one cross-attention followed by one self-attention, which can be represented as
\begin{equation}
  \label{eq_per}
  Z = f_{\textsc{per}}(X_c),
\end{equation}
where $Z \in \mathbb{R}^{n \times t'' \times d}$ is a latent vector with $t''$ number of latents ($t'' << t'$) and $d$ is latent dimension. Perceiver with an input neural ODE can have lower number of latents but Perceiver needs latents equal to sequence length to have output ODE because it needs to mask the input. Perceiver outputs a set of discrete hidden representations, as shown in Fig.~\ref{fig_model}.

\subsection{Output Neural ODE and Classifier}
\label{subsec_ode_out}
Similar to input ODE, output ODE helps to learn the time continuous dynamics of the hidden state of Perceiver and is also helpful to deal with irregular time-series and can be used to predict future risk of patient deterioration. Thus, COPER provides a double engine to handle ITS. It can be represented as:
\begin{equation}
  \label{eq_ode2}
  Y_c = f_{\textsc{ode-out}}(Z),
\end{equation}
where $Y_c \in \mathbb{R}^{n \times t'' \times d}$ represents continuous hidden state.
% \subsection{Classifier}
% \label{subsec_classifier}
Classifier layer is task specific and in this case it is a linear layer with one neuron to predict in-hospital mortality task. 
% But, COPER can be used to solve other tasks related with time-series.

\subsection{Difference from Existing Work}
\label{subsec_diff_literature}
Our work has some similarity to \cite{rubanova2019latent} and \cite{jaegle2021Perceiver}. However, our work employs neural ODEs for the continuity of input as well as output, unlike \cite{rubanova2019latent} which applies to only hidden state of RNNs. Moreover, \cite{rubanova2019latent} used encoder-decoder architecture with an ODE-RNN as encoder and ODE as a decoder. We have adapted the idea of Perceiver \cite{jaegle2021Perceiver} to time-series and we use single cross-attention followed by one self-attentions, which require use of masking to avoid leaking future information. Moreover, unlike \cite{jaegle2021Perceiver}, we do not repeat cross- and self-attention blocks with the same input, which helps to reduce complexity, but still learns well.

\section{Results}
\label{sec_results}
\subsection{Experimental Settings}
\label{subsec_exp_settings}
All the experiments are implemented in Pytorch and executed on an Ubuntu machine (64GB RAM, 1 NVIDIA GeForce GPU). Parameters of COPER are selected using trial and error as: 32 neurons in single embedding layer, 3 hidden layers of 128 neurons for each neural ODE, 128 cross- and self-attend head dimension, 64 latent dimension, dropout of 0.5 for attentions, neural ODE network as well as for LSTM. LSTM has two layers with hidden state size 50. We have used Adam optimizer with a constant learning rate of 0.0001. In addition to dropout, we use early stopping with a patience of 10 epochs to avoid overfitting. All experiments are executed three times with different seed values. The code to reproduce the results is publicly released at \url{https://github.com/jmdvinodjmd/COPER}.

\subsection{Dataset, Task and Baselines}
\label{subsec_exp_dataset}
We used the publicly available MIMIC-III dataset \cite{johnson2016mimic} for an in-hospital mortality prediction task, which contains time-series data in the ICU setting. Following \cite{harutyunyan2019multitask}, we use a dataset with 76 feature variables and 14,681, 3,236 and 3,222 samples in train, validation and test datasets. In-hospital mortality is a binary classification task to predict from the first 48 hours of ICU admission for hourly data if patient will die in the hospital or not. Mortality prediction is very important for triage, initial risk assessment, resource management and designing effective treatment plans \cite{tan2020data}. We take LSTM and Perceiver as baselines to compare with COPER at 0\%, 25\%, 50\% and 75\% removal of time steps. Since in real-life, there could be large gaps between hospital visits so we have removed the data in chunks, and to keep experiments simple, we removed in three chunks, i.e., after first observation, from middle and from end. For baselines, we employ a carry-forward technique for the missing steps, while COPER uses neural ODEs to calculate the missing steps.

\subsection{Results}
% \subsection{Convergence of COPER}
% \label{subsec_exp_convergence}
% Fig.~\ref{fig_convergence} presents the convergence of COPER using train and validation loss against number of epochs, at removal of 75\% of the time steps. The figure shows good convergence without much overfitting of the model.
% \begin{figure}[htb!]
%   \centering
%   \includegraphics[width=0.3\textwidth]{loss.png}
%   \caption{Convergence of COPER at 75\% time steps removed}
%   \label{fig_convergence}
% \end{figure}

\begin{figure}[t]
  \centering
  \includegraphics[scale=0.35]{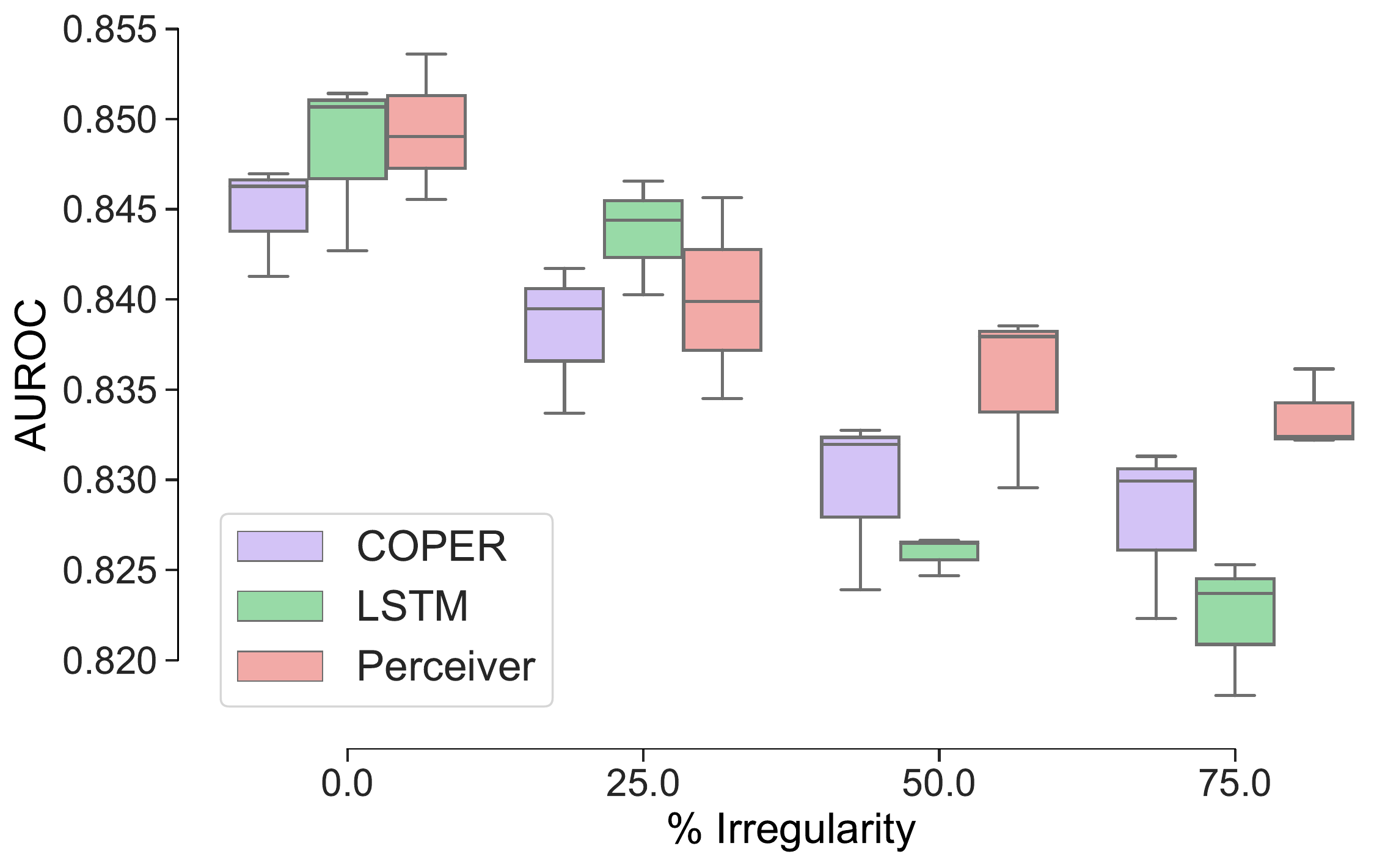}
  \caption{Comparison of COPER with baselines -- COPER copes comparatively with irregularity.}
  \label{fig_results}
\end{figure}

Fig.~\ref{fig_results}, compares COPER against the baselines, i.e., LSTM and Perceiver using AUROC by removing 0\%, 25\%, 50\% and 75\% time steps. From the results, we can make following inferences. First, comparing the models at 0\% removal of time-steps,i.e., at normal training, we observed that Perceiver outperforms others and thus can provide another alternative to recurrent networks for processing time-series data. COPER performed lesser than the Perceiver, this could be because of the difficulty in the optimization (which will be explored in the future work) as two neural ODEs are attached to Perceiver in COPER. It is noted that LSTM outperforms other models at 25\% removal of time-steps but as more time-steps are removed, it performs the worst. Interestingly, Perceiver with carry-forward handles the irregularity quite well at 50\% and 75\%. This could be because of the fact that it is a powerful model and carry-forward performs quite well in healthcare setting to deal with missingness \cite{scherpf2019predicting}. Finally, we observed that COPER performs comparatively with Perceiver at 25\% irregularity and has lesser percentage drop, also COPER outperformed LSTM at 50\% and 75\% irregularity.

Thus, the experimental evaluation shows that the Perceiver could be an alternative to recurrent networks, and COPER copes comparatively with ITS in healthcare.

% LSTM and Perceiver models also cope well with the irregularity. This is because we did not drop the time steps rather replaced with previous value, and carry-forward copes to some extent with irregularity \cite{scherpf2019predicting} due to high missing values in MIMIC-III \cite{sun2020review} and having weak dynamics.

\section{Conclusion and Future Scope}
\label{sec_conclusion}
In this paper, we used embedding and neural ODE to learn continuous patient states, and adapted Perceiver model for time series to develop a model, called COPER, to cope with irregular time-series in healthcare. COPER has continuous input and output space which helps to solve the irregularity issue, and from the Perceiver, it inherits the capability to deal with multi-modality large-scale long context inputs.

The continuity of patient state is helpful to calculate the patient's status at times when patient measurements are not available. Similarly, continuous hidden state of Perceiver allows COPER to make predictions in continuous time irrespective of the input sampling frequency. The limitation of this work is that we have evaluated the proposed framework on only MIMIC-III. In future, we will do extensive experimentation using more datasets for more rigorous performance evaluation.

% needs more computations due to two neural ODEs and it also needs detailed analysis of performance. In future, we intend to perform extensive experiments, with more tasks and more datasets to evaluate the performance, to justify use of double continuity, and thorough hyperparameter tuning will be performed. We will also study patient state dynamics for disease progression modelling. Experiments with large-scale long context inputs are also needed to justify the utility of Perceiver based architecture.

% \section*{Acknowledgment}
% The views expressed are those of the authors and not necessarily those of the NHS, the NIHR, or the Department of Health.

\bibliographystyle{IEEEtran}
\bibliography{ML4HC}

\end{document}